\documentclass[sigconf, nonacm]{acmart}
\pdfoutput=1
\newcommand\vldbyear{2026}
\newcommand\vldbworkshop{DASHSys: Systems for Data-centric Agents with Human-in-the-loop}
\newcommand\vldbauthors{\authors}
\newcommand\vldbtitle{\shorttitle}
\newcommand\vldbavailabilityurl{}
\newcommand\vldbpagestyle{empty}
\begin{document}

\title{GUIDE: Governed Unified Intelligence for Document-to-Artifact Generation in Enterprise Settings}

\author{Shivali Dalmia}
\authornote{Equal contribution}
\affiliation{%
  \institution{Centific Research}
  \city{Washington}
  \country{USA}}
\email{shivali.dalmia@centific.com}

\author{Sumukha Thoppanahalli}
\authornotemark[1]
\affiliation{%
  \institution{Centific Research}
  \city{Washington}
  \country{USA}}
\email{sumukhasharma.t@centific.com}

\author{Mohammadreza Sediqin}
\authornotemark[1]
\affiliation{%
  \institution{Centific Research}
  \city{New York}
  \country{USA}}
\email{mohammadreza.s@centific.com}

\author{Abhishek Mukherji}
\affiliation{%
  \institution{Centific Research}
  \city{Washington}
  \country{USA}}
\email{abhishek.mukherji@centific.com}

\begin{abstract}
Enterprise guideline documents are heterogeneous and multimodal, combining narrative text, complex tables, and embedded images. Existing LLM and VLM systems face hallucinated content, table structure degradation, and lack governed workflows extending beyond extraction to validation and artifact generation. This leaves enterprises to perform this manually, consuming 2–3 days per document. To address this, we introduce GUIDE, a governed multi-agent framework built on a shared versioned rule store with schema-validated inter-agent contracts and end-to-end provenance tracking. Six specialized agents handle parsing, VLM-driven extraction, consistency checking, evaluation, human-in-the-loop (HITL) escalation, and persona-tailored artifact synthesis. Evaluated on 120 real-world enterprise guideline documents, GUIDE achieves 96\% document success, extracts 3,896 rules with 71.4\% auto-approved, produces 812 deployment-ready artifacts, and reduces turnaround to 40--125 minutes per document.
\end{abstract}

\maketitle
\pagestyle{\vldbpagestyle}
\begingroup\small\noindent\raggedright\textbf{VLDB Workshop Reference Format:}\\
\vldbauthors. \vldbtitle. VLDB \vldbyear\ Workshop: \vldbworkshop.\\
\endgroup
\begingroup
\renewcommand\thefootnote{}\footnote{\noindent
This work is licensed under the Creative Commons BY-NC-ND 4.0 International License. Visit \url{https://creativecommons.org/licenses/by-nc-nd/4.0/} to view a copy of this license. For any use beyond those covered by this license, obtain permission by emailing \href{mailto:info@vldb.org}{info@vldb.org}. Copyright is held by the owner/author(s). Publication rights licensed to the VLDB Endowment. \\
\raggedright Proceedings of the VLDB Endowment. %, Vol. \vldbvolume, No. \vldbissue\ %
ISSN 2150-8097. \\
}\addtocounter{footnote}{-1}\endgroup

\ifdefempty{\vldbavailabilityurl}{}{
\vspace{.3cm}
\begingroup\small\noindent\raggedright\textbf{VLDB Workshop Artifact Availability:}\\
The source code, data, and/or other artifacts have been made available at \url{\vldbavailabilityurl}.
\endgroup
}




\section{Introduction}

Enterprise annotation pipelines rely on unstructured guideline documents that must be converted into structured, executable work artifacts before labeling can begin. Quality managers (QMs) and project managers (PMs) currently perform this manually---reading guidelines, inferring rules, resolving ambiguities, and assembling annotator instructions and statements of work. Each document takes 2--3 days, is error-prone, and must be redone whenever documents are updated~\cite{anderson2024design, perot2024lmdx}. At scale, this becomes a critical bottleneck limiting how quickly projects can be staffed, launched, and maintained.

At its core, this is a data management challenge. Rules extracted from heterogeneous documents must be versioned, deduplicated, validated against a schema, and made traceable from source to artifact. Enterprise guideline documents compound this: text is encoded as positional tokens rather than semantic units; tables span pages, contain merged cells, or appear as images; and visual elements such as annotation examples and bounding box diagrams carry semantically critical information not captured in text~\cite{ke2025large}. Any parsing error cascades into downstream rule quality.

We introduce GUIDE, a governed multi-agent framework that treats guideline-to-artifact conversion as a data management problem: six specialized agents coordinate through a shared staging store that holds all extracted rules and intermediate results as structured, versioned data. This enforces contractual guarantees between stages, makes every artifact traceable to its source, and triggers human review only at explicit thresholds, the properties that make the pipeline governed. Our contributions are:
\begin{itemize}
  \item \textbf{A versioned, schema-validated shared staging store} with stable \texttt{rule\_id} keying, inter-agent schema contracts, and end-to-end provenance tracking, forming the data management backbone of the pipeline.
  \item \textbf{GUIDE}, a governed multi-agent framework integrating deterministic parsing, VLM-based extraction, structured rule modeling, and dependency-aware artifact generation across six specialized agents.
  \item \textbf{A two-stage evaluation framework} combining structural validation (L1) and LLM-based semantic scoring (L2) with automated acceptance, targeted regeneration, and selective HITL escalation; zero-edit approvals feed back as calibration signals, reducing reviewer load.
  \item \textbf{A dependency-driven HITL workflow} routing only flagged rules, gaps, and artifacts to the QM or PM workbench by object type, severity, and downstream dependency.
  \item \textbf{Empirical evaluation} on 120 real-world enterprise guideline documents demonstrating strong performance across extraction fidelity, consistency control, and persona-specific artifact generation.
\end{itemize}

\section{Related Work}

\textbf{Document Parsing and Extraction.}
Document parsing has evolved from rule-based OCR pipelines~\cite{smith2007overview} to hybrid vision-language approaches~\cite{poznanski2025olmocr, li2025monkeyocr}, alongside
discourse-level segmentation and structure-aware
methods~\cite{LACES-NLPIR, sediqin2025rst}. Table extraction remains difficult: rule-based methods rely on geometric heuristics while VLM-based approaches improve robustness but introduce instability on dense or borderless layouts~\cite{verbovskiy2025comparing, kim2022donut}. Recent VLMs like Qwen-VL~\cite{wang2024qwen2, bai2025qwen3} and LLaVA~\cite{liu2023visual} advance multimodal extraction but vary in hallucination and visual grounding~\cite{zhu2024mmdocbenchbenchmarkinglargevisionlanguage}.

\textbf{LLM-Based Structured Extraction.}
GoLLIE~\cite{sainz2024gollieannotationguidelinesimprove} shows annotation guidelines in prompts improve zero-shot extraction; UIE~\cite{lu2022unifiedstructuregenerationuniversal} enhances robustness via instruction tuning. However, these methods treat extraction as a single-step process and lack validation, contradiction handling, or production-level structuring. Layout-aware parsers such as Docling~\cite{livathinos2025docling} and Donut~\cite{kim2022donut} focus on extraction as a terminal task with no governed output routing or artifact generation. Since their outputs are structurally incompatible with GUIDE's
schema-constrained rule format, a direct end-to-end comparison is
ill-posed; we instead compare against a monolithic single-model baseline
with identical input and output structure.

\textbf{Multi-Agent and HITL Systems.}
Multi-agent frameworks~\cite{autogen2023} decompose complex tasks across specialized agents but coordinate via unstructured message passing, without schema enforcement or provenance guarantees. HITL systems  improve annotation quality through selective routing~\cite{amershi2016human}, yet treat review as a flat queue without distinction by object type, severity, or dependency. GUIDE addresses these gaps via schema-enforced inter-agent contracts, multi-stage quality evaluation, and dependency-aware HITL escalation.

\section{System Architecture}
\label{sec:arch}

\begin{figure}[t]
\centering
\includegraphics[width=\linewidth]{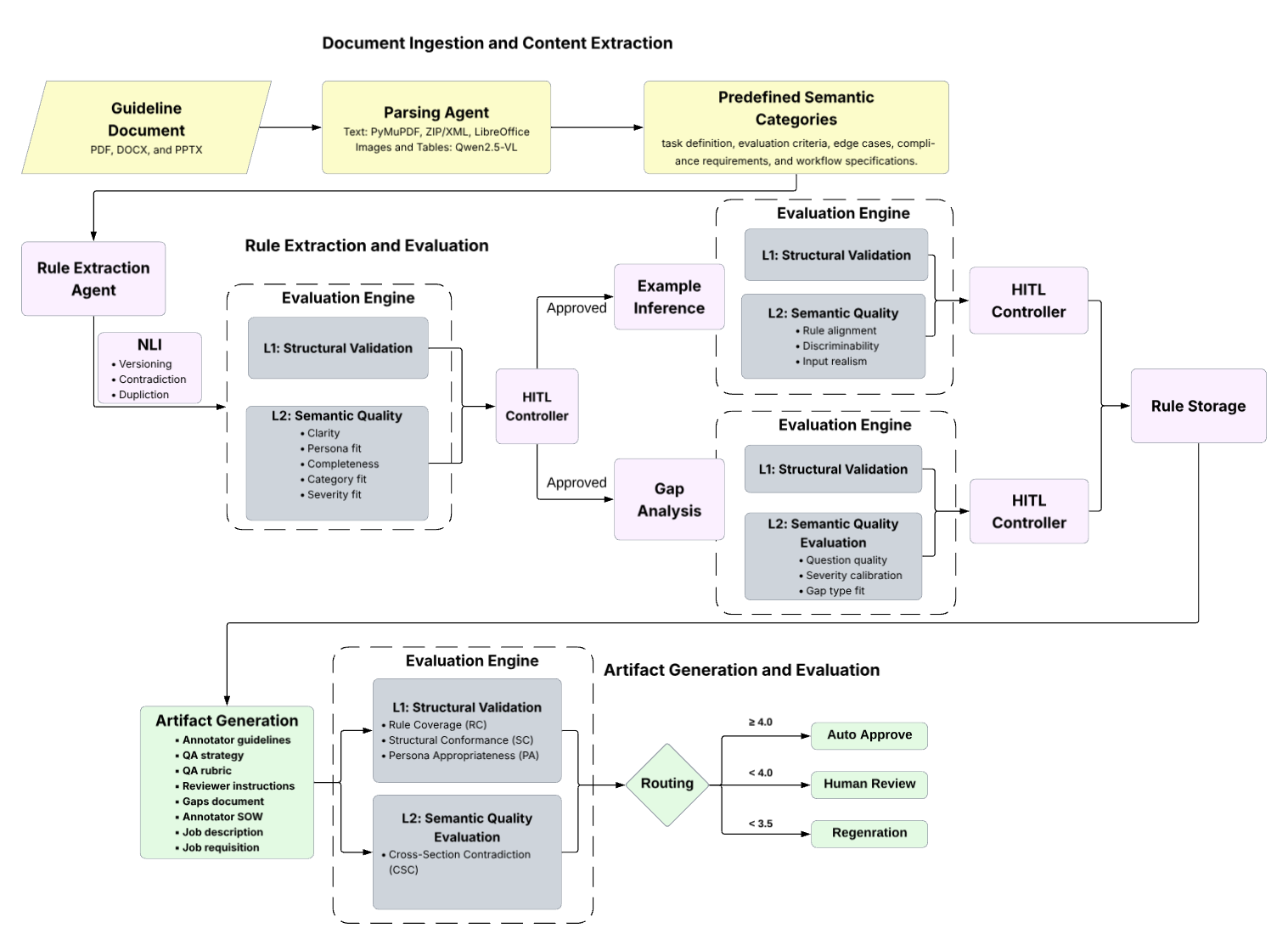}
\caption{Overview of the GUIDE pipeline.}
\label{fig:pipeline_overview}
\end{figure}

\textsc{GUIDE} is architected around a central versioned rule store: schema-enforced tables keyed by stable \texttt{rule\_id}s serving as the shared data layer. Agents read from and write to this store through Pydantic-validated contracts, ensuring no downstream agent ever consumes structurally invalid data. This yields three properties essential for enterprise deployment: \textit{provenance} (every artifact traces to its source rules and originating document), \textit{versioning} (rule updates across document revisions are reconciled rather than reprocessed), and \textit{auditability} (every HITL decision is logged against a stable identifier). The system comprises six components: Parsing Agent, Rule Extraction Agent, Consistency Module, Evaluation Module, HITL Controller, and Artifact Generation Agent (Figure~\ref{fig:pipeline_overview}).

\subsection{Document Ingestion and Rule Extraction}

The Parsing Agent separates deterministic text extraction from VLM processing. Text is extracted from PDF (PyMuPDF), DOCX (internal XML), and PPTX (LibreOffice $\to$ PDF) without language models. Visual content is processed using Qwen2.5-VL~\cite{wang2024qwen2} after MD5-based image deduplication. Extraction quality is assessed via $Q = 1 - (\text{garbage} + \text{mojibake} + \text{repetition} + \text{silent\_skip})$ where each term is a normalized defect rate. Extracted content is segmented into semantic categories (task definition, evaluation criteria, edge cases, compliance requirements, workflow specifications) as a routing layer for downstream processing.

The Rule Extraction Agent applies a two-stage pipeline: open-domain extraction identifying candidate rules with source spans and confidence scores, followed by normalization into a fixed 26-field schema. Rule type determines persona routing: \texttt{evaluation-criteria}, \texttt{edge-case}, and \texttt{qa-process} rules route to the QM workbench; \texttt{worker-requirements} and \texttt{delivery-schema} rules route to the PM workbench. Rules pass through the Consistency Module, which applies embedding-based similarity filtering followed by NLI classification~\cite{maccartney2009natural} for deduplication and version alignment. Gap analysis identifies missing or ambiguous aspects as structured \textit{GapObject}s; resolved gaps generate \textit{ClarificationRecord}s and new \textit{RuleUnit}s. For each approved rule, examples are extracted when available or inferred under strict adherence to rule semantics.

\subsection{Evaluation Engine and HITL}
As ground truth annotations are unavailable for this corpus, all evaluation
metrics are computed using standard signals: cosine similarity for
grounding, schema validation for structural compliance, and LLM-as-judge
scoring for semantic quality. All thresholds were selected empirically over
the full corpus and fixed prior to all reported experiments; with only 120
diverse production documents, a held-out split would have reduced the
diversity available for calibration. To validate the LLM judge, 300
rule-level annotations were independently labeled by expert annotators
blind to judge outputs; the judge achieves precision 0.941, recall 0.974,
F1 0.957, and Cohen's $\kappa = 0.813$ against these labels, providing an
independent check that the calibrated thresholds align with expert judgment
rather than overfitting the calibration corpus. These metrics and
thresholds are applied through a two-stage validation pipeline that every
structured object must pass. \textbf{L1 (Structural)} applies deterministic
Pydantic validation: 28 constraints for \textit{RuleUnit} (required fields,
valid enumerations, prohibition of instruction-source duplication), 4 for
\textit{ExampleObject} (valid rule linkage, separation of correct/incorrect
outputs), and 8 for \textit{GapObject} (valid gap types, resolved rule
references). Failed objects are retried, corrected, or rejected by error
severity.

\textbf{L2 (Semantic)} scores passing objects across $K$ quality dimensions using an LLM-as-judge:
\begin{align}
   S(x) = \frac{1}{K} \sum_{k=1}^{K} s_k(x), \quad s_k \in \{1,\ldots,5\}
\end{align}
with $K{=}5$ for \textit{RuleUnit} (clarity, persona fit, completeness, category fit, severity fit), $K{=}3$ for \textit{ExampleObject} (rule alignment, discriminability, input realism), and $K{=}3$ for \textit{GapObject} (question quality, severity calibration, gap type fit). Routing is by minimum dimension score: $\min_k s_k \geq 4$ auto-approves; any $s_k \in \{2,3\}$ routes to HITL; any $s_k = 1$ rejects. Objects with \texttt{rule\_source = inferred} always route to HITL. Zero-edit HITL approvals are logged as calibration data for L2, progressively reducing reviewer load over deployment cycles. Every review decision is recorded against the corresponding rule\_id, providing a complete audit trail from human action to source rule.

HITL is staged and dependency-aware: Phase 1 reviews \textit{RuleUnit}s; Phase 2 reviews \textit{GapObject}s conditioned on the approved rule set; Phase 3 reviews \textit{ExampleObject}s conditioned on finalized rules and resolved gaps. This logically bounds the downstream review surface and ensures QM effort focuses on outputs grounded in high-quality, approved rules.

\subsection{Artifact Generation}

The Artifact Generation Agent transforms approved rules, resolved gaps, and evaluated examples into eight deployment-ready artifacts via Pydantic-constrained templates: annotator guidelines, QA strategy, QA rubric, reviewer instructions, gaps document, QA agent specification~\cite{kothari2026position}, annotator SOW, and job description/requisition. Each artifact is evaluated via:
\begin{align}
\text{ART} = w_1 \cdot \text{RC} + w_2 \cdot \text{SC} + w_3 \cdot \text{PA} + w_4 \cdot \text{CSC}
\end{align}
where RC (rule coverage), SC (structural conformance), PA (persona appropriateness via Flesch readability and LLM judgment), and CSC (cross-section contradiction, evaluated by Claude Sonnet 4.6) are combined with empirically tuned weights. $\text{ART} \geq 4.0$ auto-approves; $3.5 \leq \text{ART} < 4.0$ routes to human review; $\text{ART} < 3.5$ triggers regeneration.

\section{Evaluation}

\subsection{Dataset and VLM Selection}

Our corpus consists of 120 enterprise guideline documents from industrial clients (67 text, 23 speech, 16 multimodal, 8 image, 6 video; PDF/DOCX/PPTX formats), used as received without preprocessing; documents are governed by client confidentiality and non-disclosure agreements and cannot be released. Complexity tiers are  defined  by modality composition: \textit{Low} (${\sim}74$ KB, 39--46 min), \textit{Moderate} (${\sim}1.5$ MB, 46--70 min), and \textit{High} (${\sim}4.2$ MB, 65--125 min). Table~\ref{tab:vlm_main} shows Qwen2.5-VL-32B achieves the best balance of evidence rate (77.8\%), hallucination (20.2\%), and throughput (232/355 images) and the lowest duplication (14.6\%); we select it for all stages.
%Table~\ref{tab:system_compare_vlm} situates GUIDE against representative baselines; GUIDE is the only system providing explicit schema-based grounding, consistency checking, full validation, and a complete governed pipeline.

\begin{table}[t]
\caption{VLM benchmark results across document extraction tasks.}
\label{tab:vlm_main}
\small
\begin{tabular}{lccccc}
\toprule
Model & Evidence & Halluc. & Quality & Thruput & Dupl. \\
      & (\%)     & (\%)    & (\%)    & (img)   & (\%)  \\
\midrule
\textbf{Qwen2.5-VL-32B} & \textbf{77.8} & \textbf{20.2} & 64.9 & \textbf{232} & \textbf{14.6} \\
Qwen3-32B      & 76.7 & 22.7 & \textbf{73.2} & 184 & 17.9 \\
LLaVA-13B      & 64.1 & 35.9 & 63.6 & 159 & 42.0 \\
\bottomrule
\end{tabular}
\end{table}

%\begin{table}[t]
%\caption{Capability comparison of %extraction systems.}
%\label{tab:system_compare_vlm}
%\centering
%\small
%\setlength{\tabcolsep}{2pt}
%\resizebox{\columnwidth}{!}{%
%\begin{tabular}{lccccc}
%\toprule
%System & Grounding & Extraction & %Consistency & Validation & Post-proc. \\
%\midrule
%Docling        & Weak (OCR)         & Yes & No   & No      & Rule-based \\
%Donut          & None               & Yes & No   & No      & None \\
%UIE            & None (prompt)      & Yes & No   & Limited & Prompt \\
%GoLLIE         & None (instruction) & Yes & Weak & Limited & Instr. tuning \\
%Qwen2.5-VL     & Implicit           & Yes & No   & No      & Minimal \\
%\textbf{GUIDE} & Explicit (schema)  & Yes & Yes  & Yes     & Full pipeline \\
%\bottomrule
%\end{tabular}
%}
%\end{table}

\subsection{Content Extraction, Rule Extraction, and Artifacts}

Content extraction across 120 documents (Table~\ref{tab:content_summary}) achieves 96\% document success with 99.2\% page coverage, 97.1\% figure recall, 88.3\% table recall, and $Q{=}1.00$ on successfully processed documents. Failures are caused by VLM timeouts on image-heavy documents and poorly structured tables.

Rule extraction on 115 documents (Table~\ref{tab:rule_summary}) yields 3,896 \textit{RuleUnit}s with 84.8\% evidence rate, 82.6\% coverage, and 3.2\% hallucination. L1 passes 99.1\%; the 0.9\% flagged are structurally valid but insufficiently precise for direct execution. L2 auto-approves 71.4\% with 28.6\% routed to HITL and 0\% rejected; most HITL cases arise from incomplete semantic coverage where rules capture the primary case but miss edge conditions. The consistency module identifies 26.7\% gaps, 3.0\% duplications, and 2.9\% contradictions. The 2--3 day manual baseline is an expert estimate by the QMs and PMs who
perform this task, indicating an order-of-magnitude reduction rather than
an exact head-to-head.

Artifact generation produces 812 artifacts (Table~\ref{tab:artifact_summary}). Cross-section contradiction (93.7\%) and structural conformance (83.1\%) reflect strong logical and structural consistency. Rule coverage (56.9\%) and persona appropriateness (59.8\%) remain the most challenging dimensions, the former reflects partial instantiation under constrained generation context, the latter reflects the difficulty of adapting technical rules for non-expert audiences. Only 29.8\% of artifacts are auto-approved, with 52.0\% routed to human review and 18.2\% rejected, revealing two primary failure modes: incomplete rule propagation and persona adaptation gaps. 
\subsection{Comparison Against a Monolithic Baseline}
\label{sec:baseline}
We compare GUIDE against a one-pass Qwen2.5-VL-32B baseline that produces
rules and artifacts directly from parsed documents, without the rule
store, L1/L2 validation, consistency module, or HITL routing, using the
same parser, documents, and scoring. Removing governance degrades every
dimension (Table~\ref{tab:baseline}): hallucination rises from 3.2\% to
15.7\%, duplication from 3.0\% to 10.3\%, contradictions from 2.9\% to
7.8\%, and L1 pass rate falls from 99.1\% to 93.2\%. A layer-wise
ablation isolates each layer: L1/L2 routes 1,114 units (28.6\%) to
review; the consistency module removes 117 duplicates, routes 113
contradictions, and surfaces 1,040 otherwise-undetected gaps.

\begin{table}[t]
\caption{Ungoverned one-pass baseline vs.\ GUIDE (same model, documents, and scoring).}
\vspace{-6pt}
\label{tab:baseline}
\centering
\footnotesize
\renewcommand{\arraystretch}{0.95}
\setlength{\tabcolsep}{8pt}
\begin{tabular}{lcc}
\toprule
Metric & Baseline & GUIDE \\
\midrule
Hallucination  & 15.7\% & \textbf{3.2\%} \\
Duplication    & 10.3\% & \textbf{3.0\%} \\
Contradiction  & 7.8\%  & \textbf{2.9\%} \\
L1 pass rate   & 93.2\% & \textbf{99.1\%} \\
\bottomrule
\end{tabular}
\end{table}

\begin{table}[t]
\small
\setlength{\tabcolsep}{1pt}
\caption{Content extraction evaluation across 120 documents.}
\vspace{-8pt}
\label{tab:content_summary}
\begin{tabular}{llc}
\toprule
Category & Metric & Value \\
\midrule
Coverage
& Page coverage     & 99.2\% \\
& Figure recall     & 97.1\% \\
& Table recall      & 88.3\% \\
\midrule
Quality score
& Overall score     & 1.00   \\
& Garbage ratio     & 0.00\% \\
& Mojibake ratio    & 0.00\% \\
& Repetition ratio  & 0.01\% \\
& Silent skip ratio & 0.00\% \\
\midrule
Success rate
& Document success  & 96\% (115/120) \\
\bottomrule
\end{tabular}
\end{table}

\begin{table}[t]
\caption{Rule extraction (3,896 RuleUnits).}
\label{tab:rule_summary}
\centering
\footnotesize
\renewcommand{\arraystretch}{0.95}
\setlength{\tabcolsep}{3pt}
\begin{tabular}{llc}
\toprule
Category    & Metric         & Value  \\
\midrule
Quality     & Evidence rate  & 84.8\% \\
            & Coverage       & 82.6\% \\
            & Hallucination  & 3.2\%  \\
\midrule
L1          & Pass rate      & 99.1\% \\
            & Ambiguity      & 0.9\%  \\
\midrule
L2          & Auto-approved  & 71.4\% \\
            & Human review   & 28.6\% \\
            & Rejected       & 0.0\%  \\
\midrule
Consistency & Gaps           & 26.7\% \\
            & Duplications   & 3.0\%  \\
            & Contradictions & 2.9\%  \\
\bottomrule
\end{tabular}
\end{table}

\begin{table}[t]
\caption{Artifact evaluation (115 documents, 812 artifacts).}
\label{tab:artifact_summary}
\centering
\footnotesize                          % was \small
\renewcommand{\arraystretch}{0.95}     % tighter rows
\setlength{\tabcolsep}{3pt}
\begin{tabular}{lcc}
\toprule
Category & Metric & Value \\
\midrule
HITL routing & Auto-approved       & 29.8\% \\
             & Human review        & 52.0\% \\
             & Rejected            & 18.2\% \\
\midrule
L1           & Rule coverage       & 56.9\% \\
             & Structural conform. & 83.1\% \\
             & Cross-sect. contra. & 93.7\% \\
             & Gap completeness    & 66.7\% \\
\midrule
L2           & Persona approp.     & 59.8\% \\
             & Sequential correct. & 65.2\% \\
\bottomrule
\end{tabular}
\end{table}

\section{Limitations and Future Work}

\textbf{Limitations.} GUIDE demonstrates strong performance across content extraction, rule generation, consistency validation, and artifact generation on real-world enterprise documents. VLM extraction stability decreases on low-quality scans and borderless or merged-cell tables, a known challenge across current vision-language systems. Persona appropriateness and rule coverage remain the most challenging artifact dimensions, reflecting the inherent difficulty of adapting technical rules to non-expert audiences. The calibration mechanism relies on accumulating zero-edit HITL approvals over deployment cycles, so scoring stability in early cycles remains limited. The current evaluation also covers English enterprise guidelines only.
While the 26-field rule schema and QM/PM persona routing are specific to
annotation operations, the staging store, schema contracts, L1/L2
evaluation, and dependency-aware HITL routing are domain-agnostic.

\textbf{Future Work.} We will improve table detection and VLM prompting for degraded layouts, develop persona adaptation modules that learn audience-specific language patterns from approved artifacts, and incorporate structured reviewer edits as additional fine-tuning signal to accelerate calibration convergence. GUIDE will further be extended to multilingual settings and broader domains including legal, clinical, and regulatory, with domain-adapted consistency models trained on enterprise rule pairs.

\section{Conclusion}

We presented GUIDE, a governed multi-agent framework that transforms heterogeneous enterprise guideline documents into structured, deployment-ready artifacts via a schema-enforced shared staging store with end-to-end provenance. By coordinating six specialized agents through typed intermediate relations and applying threshold-driven HITL escalation, GUIDE reduces end-to-end turnaround from 2--3 days to 40--125 minutes while maintaining strong extraction fidelity and consistency. GUIDE shows how governed multi-agent pipelines serve as a principled foundation for data-aware, human-aligned agentic systems where reliability, traceability, and selective human oversight are first-class design goals.
\begin{acks}
We thank Srinivasa Karthikeya Reddy Kovvuri for his contributions to this work.
\end{acks}

\bibliographystyle{ACM-Reference-Format}
\bibliography{reference}

\end{document}